\documentclass[journal,comsoc,onecolumn]{IEEEtran}
\usepackage[T1]{fontenc}
\usepackage{cite}
\usepackage{amsmath,amssymb,amsfonts}
\usepackage{algorithmic}
\usepackage{graphicx}
\usepackage{textcomp}
\usepackage{xcolor}
\usepackage{tikz}
\usepackage{units}
\usepackage{float}
\usepackage{lipsum,graphicx,multicol,subcaption}
\def\BibTeX{{\rm B\kern-.05em{\sc i\kern-.025em b}\kern-.08em
    T\kern-.1667em\lower.7ex\hbox{E}\kern-.125emX}}
\begin{document}
\pdfoutput=1
\title{Slider: On the Design and Modeling of a 2D Floating Satellite Platform}

\author{{Avijit Banerjee, Jakub Haluska, Sumeet G. Satpute, Dariusz Kominiak, and George Nikolakopoulos\IEEEauthorrefmark{1}}
\thanks{\IEEEauthorrefmark{1} Robotics and Artificial Intelligence, Department of Computer, Electrical and
Space Engineering, Lule\aa\,\,University of Technology, Lule\aa\,\, 
        {\tt\small\{aviban, jakhal, sumsat, darkom,  geonik\}@ltu.se}}%
}
\maketitle
\begin{abstract}
In this article, a floating robotic emulation platform for a virtual demonstration of satellite motion in space is presented. The robotic platform design is characterized by its friction-less, levitating, yet planar motion over a hyper-smooth surface. The robotic platform, integrated with sensor and actuator units, is fully designed and manufactured from the Robotics and Artificial Intelligence Team at Lule\aa\ University of Technology. A detailed design description  along with the mathematical modeling describing the platform's dynamic motion is formulated. Finally, the proposed design is validated in extensive simulation studies, while the overall test bed experimental setup, as well as the vehicle hardware and software architectures, are discussed in detail. Furthermore, the entire design, including 3D printing CAD model and different testbed elements, is provided in an open-source repository and a test campaign is used to showcase its capabilities and illustrate its operations. \end{abstract}
\IEEEpeerreviewmaketitle
\section{Introduction}
The advancements of space technology in the recent era has revolutionised our perception about space activities. Space agencies across the globe focus on the autonomous robotic missions that enable on-orbit servicing, manufacturing and maintenance of satellites, close monitoring, docking and active debris removal. Since the past decade with the growing interest of small scale satellites \cite{toorian2008cubesat}, the present-generation space missions focus on the prospect of multiple spacecraft formation missions towards autonomous operations in space. Such ambitious tasks inherit highly complex and challenging objectives that require efficient and effective autonomous guidance, navigation and control systems to ensure the overall mission success \cite{sorgenfrei2014operational}. 

To autonomously carry out such highly complex widespread objectives, advanced navigation, guidance, and control (GNC) techniques are of primary importance \cite{sorgenfrei2014operational} in executing these challenging operations with a high degree of reliability and accuracy. To ensure higher technology readiness levels before deploying into space, rigorous validation of the robotic space system through ground-based high fidelity campaigns in the relevant condition is essential for cost-effective, low-risk, and potentially high return solutions. 

In view of that, a friction-less, micro gravity orbital motion is one of most the critical aspects of the space environment, that needs to be replicated in hardware-in-the-loop tests \cite{bodin2009prisma}. However, in reality, recreating the space like micro-gravity condition is indeed challenging to set up in laboratory conditions on the Earth. A comprehensive study reported in \cite{historical} provides a systematic review on development of such simulating environment. Performing a parabolic flights test \cite{parabolic} with a free fall maneuver for small period is presented in \cite{parabolic} as potential micro-gravity research tool. Another possibility is to conduct a drop-towers test \cite{droptower}, which can provide realistic micro-gravity condition. A parabolic flight test based experimental validation related to space robotics and on orbit maintenance were published in \cite{menon2005free}. These techniques, however, are constrained by the overall flight time and limited space that could be provided inside dropped capsule \cite{air_bearing}. Moreover, such approaches  are significantly expensive for simulating the micro-gravity environment.

Researchers have suggested innovative methods for long duration flight tests for space robotic emulation. One such approach considers an  underwater natural bouncy system to replicate weightlessness \cite{underwater}. The underwater test facility provides an significant capacity for an astronaut's training, the utility of submerging a satellite is significantly limited. Another approach considers a weight reducing suspension system \cite{brown1994novel} to counter balance the gravitational force. The concept has been validated with a robotic manipulator. However, the limited allowable motion and disturbances introduced by suspension mechanism restricts the applicability of such method.  

In view of generating an approximate micro-gravity environment, planar air-bearing based mechanisms provide the most flexible dynamic equivalency with ideal space representative framework. Robotic vehicles supported by air bearing, representing spacecrafts/satellites, have some level of manoeuvring capability to move over a smooth planar nearly friction-less surrounding. Air-bearings attached with the platform releases pressurized air and creates a thin film to levitate the platform, and thereby counter balance its weight to produce a micro-gravity effect (in-plane components of gravity on the test vehicles are negligible). Thus emulating the drag free and weightless environment of orbital spaceflight. The only limitation of the design is that they are restricted to three degrees of freedom, i.e., two translation and one rotation motion, which is also closely resembles with a space scenario (since out of plane motion are very limited for actual space mission). Moreover, these facilities provide various hardware phenomena (e.g. realistic actuation mechanism, computational constraints, sensor noise, actuator uncertainty, delay etc.). In this manner, an air bearing based friction less platform provides GNC testbeds for rigorous validation, both in terms of software evaluation, as well as hardware-based implementation in high fidelity test environments that have the capability to emulate realistic conditions in space. Towards this direction, it should be mentioned that up to best of our knowledge, there are no such commercially available platforms for these friction-less micro gravity emulation platform. Various government organization and university laboratories across the globe has indigenously constructed their own test bed facilities. Examples can be found in \cite{sabatini2012design}, \cite{zappulla2017dynamic}, \cite{gallardo2011advances}. Primarily these emulation platform synthesises planner motion of a robotic vehicle, while in some designs, additional degrees of freedom are achieved by adding an air-bearing on top of the planar platform \cite{5dof, 6dof, 6dof2}. It should be noted that there is usually no thorough characterization of the test beds in the literature, which restricts a through comparison of the experimental results obtained by using various test facilities. 

In line with the presented background, this article aims in presenting the design of a novel floating platform, the Slider named from now and on, which has been fully designed and manufactured developed from the Robotics and Artificial Intelligence Team \cite{RAIWebpage} at Lule\aa\ University of Technology in Sweden. Figure~\ref{Kiruna_Facility_Frictionless_Table} depicts the hardware-in-loop test-bed facility, which consists of an epoxy-topped flat table, with a robotic arm connected to a two-dimensional gantry system and the slider platform on top of the flat table, and an ABB industrial robot in the vicinity of the flat table. The platform will be a useful tool to advance the state of the art of GNC evaluation and can be used to perform end-to-end system-level verification and validation before the system's operational deployment. 

The main contributions of the article stems from: a) the introduction of a novel design of a planar floating platform, Slider, with a detailed design specification of the floating platform based on a limited number of air bearings, b) a full mathematical model derivation for representing the transnational and rotational motion of the slider platform over the friction-less table that is required for the sequential control design step, and c) on top of that an analytical mathematical model of the framework suitable for control design is established, while also presenting a low level actuator design framework, specifically designed for the proposed platform enabling an accurate actuation. The entire design of our slider platform including 3D CAD model, data for 3D printing, laser cutting, blueprint diagram and an extensive list of various components are made available in the GitHub repository \cite{github}. We sincerely believe that our design's open-sourcing will befit interested space research communities to rebuild the hardware platform in an individually customized setup quickly. A visual demonstration of the slider in operational mode can be found in \cite{Youtube}.   

The rest of the article is structured as follows. In Section \ref{Description}, a design description of the physical floating platform equipped with various components is presented. An initial validation of the overall design is described in Section \ref{initial_test}. In Section \ref{Mathematical_model}, the mathematical model of the floating platform is formulated. An open loop simulation with low level actuator selection logic has been developed and the mathematical model has been validated numerical simulations in Section \ref{Simulation_Results} and finally, the article is concluded with its future direction in Section \ref{Conclusion}. 

\section{Design Description of Slider Platform} \label{Description}
\begin{figure}[h]
\centerline{\includegraphics [width=0.8\textwidth]{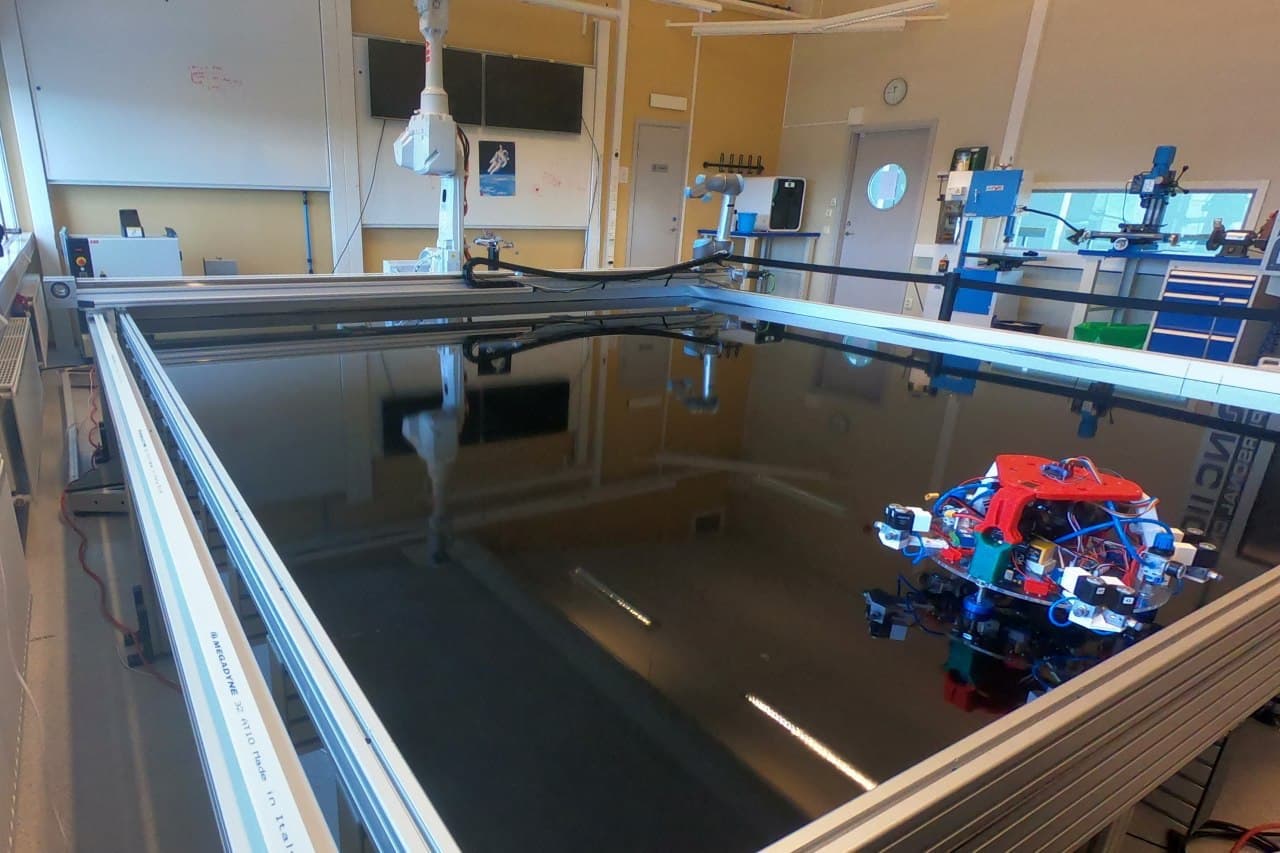}} 
\caption{Hardware-in-loop testing facility with a $4\times 4~\unit{m}$ epoxy-topped flat table, floating platform (Slider), and two robotic manipulators at Lule\aa University of Technology. }\label{Kiruna_Facility_Frictionless_Table}
\end{figure}

\begin{figure} 
\begin{subfigure}{.5\textwidth}
  \centering
  \label{Schematic_Platform_Model_Side_view}
      \includegraphics[width=1\textwidth]{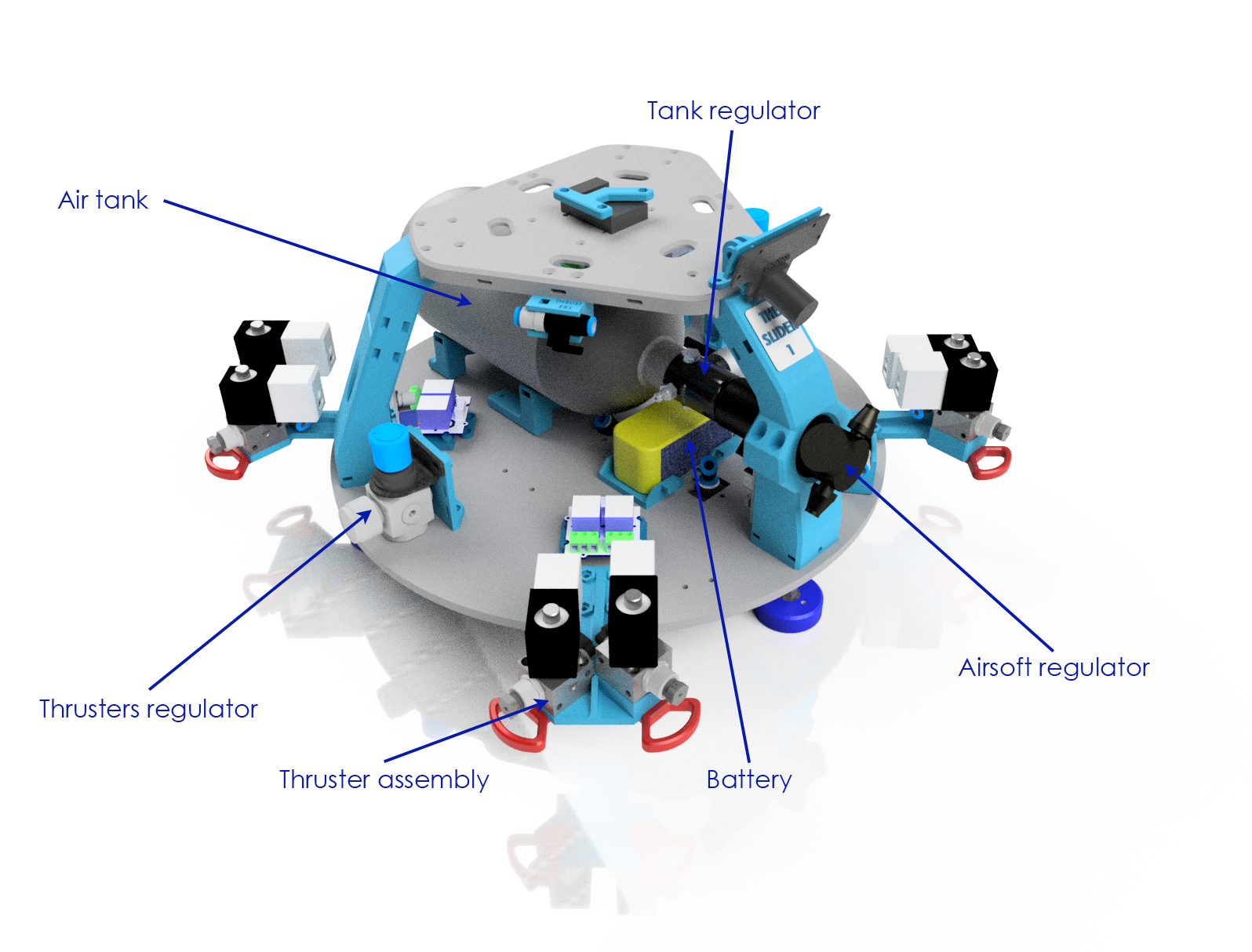}
      \caption{Side View}
\end{subfigure}%
\begin{subfigure}{.5\textwidth}
  \centering
    \includegraphics[width=1\textwidth]{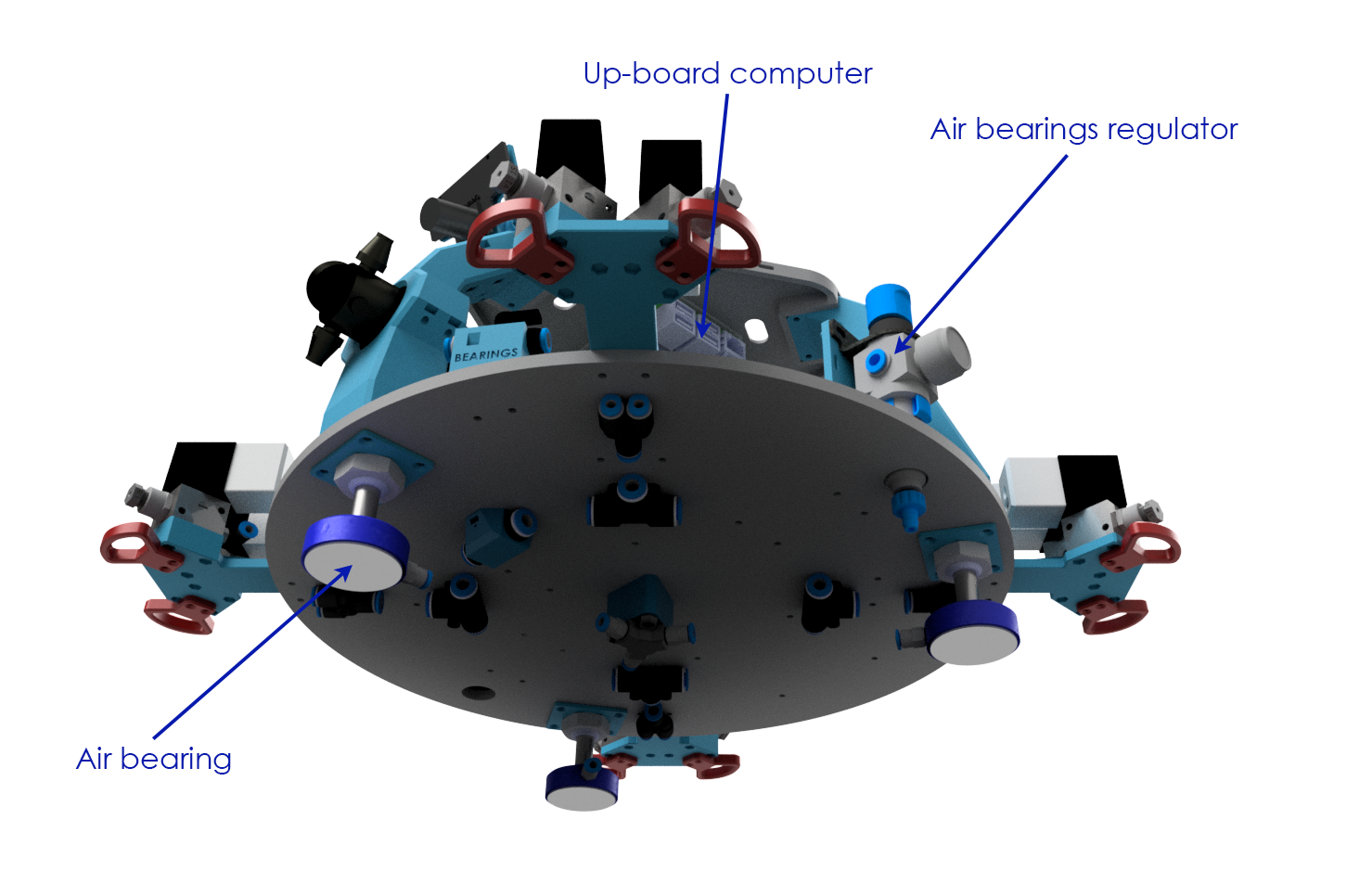}
    \caption{Bottom View}
\end{subfigure}
 \caption{Physical model of slider platform}
 \label{Schematic_Platform_Model}
\end{figure}

The robotic emulator platform is designed to smoothly maneuver over a friction-less table. The flat top surface of the table (shown in Fig. \ref{Kiruna_Facility_Frictionless_Table}) is coated with epoxy resins which creates a smooth and flat table surface required to replicate the friction-less motion of a spacecraft in space environment. A schematic design of the slider platform is presented in Fig. \ref{Schematic_Platform_Model}. The slider is supported with three air-bearings attached at its bottom deck. The functional surface of each air-bearings is porous in nature. Compressed air is evenly released through these small holes, which eventually creates an air cushion. The air-cushion supports the weight of the slider platform and allow it to levitate of the epoxy-topped table. Indeed, it can not offer a micro-gravity framework, however, such a mechanism provides a nearly friction-less environment along the 2-dimensional plane of the flat-table, which closely resembles a motion in space. Due to this fact, it is preferred as an emulation platform to demonstrate the state-of-the-art autonomous technologies for complicated space missions. The slider platform is consisting of the various subsystems described as follows

\begin{figure} 
\centerline{\includegraphics [width=0.8\textwidth]{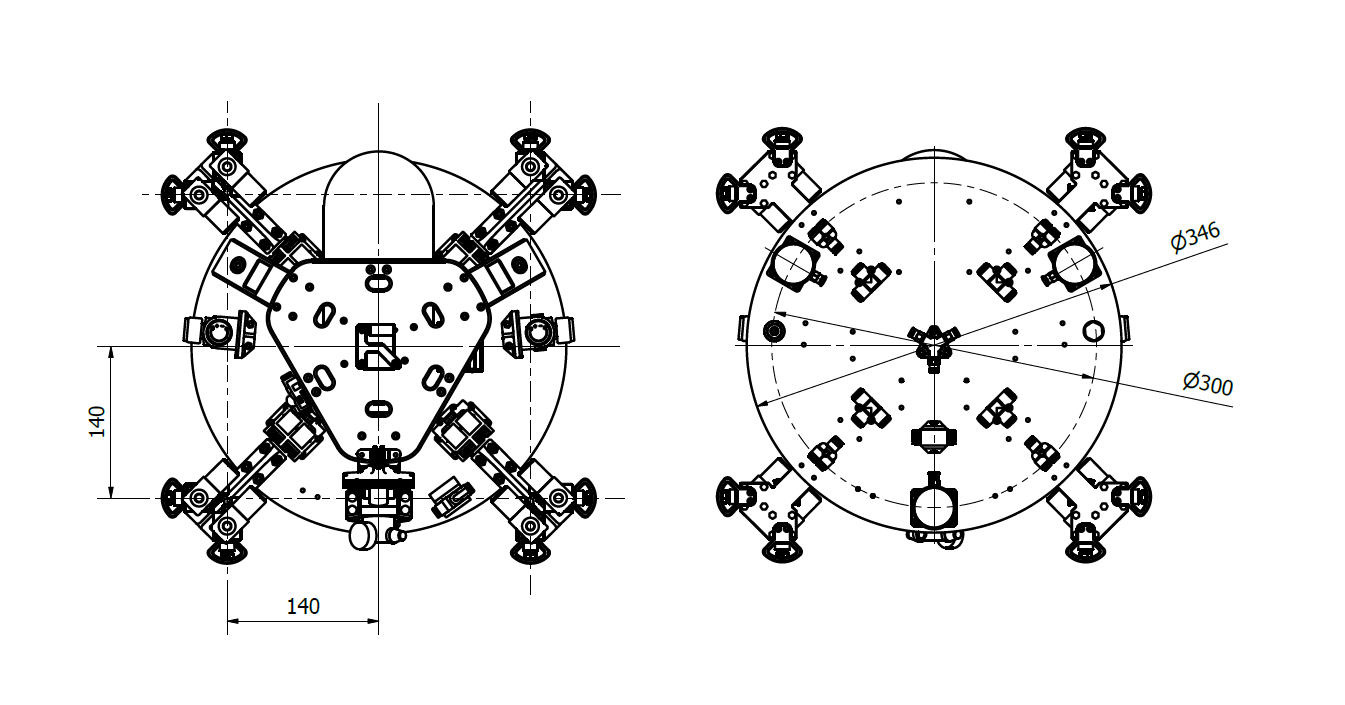}} 
\caption{Blueprint model of slider platform}
\label{Blueprintmodel}
\end{figure}

\subsection{Structural Design of Slider}
  The structural design of the platform is constructed in such way that it is light-weight, supports the necessary payload components (e.g. air bearings, compressed air tank, thrusters etc.) and provide sufficient rigidity to the over all assembly. One of the key features of the design is the easiness of manufacturing. The technologies used for production are primarily 3D printing and laser cutting. The physical construction of the outline structure is consist of a circular base along with a ceiling surface, which are connected by three `top-down-frames' as shown in Fig.\ref{Schematic_Platform_Model}. The circular base of the structure is made out of $6mm$ poly-carbonate sheet, which is light-weight and has sufficient stiffness to hold various components mounted over it. The substantial components, which  and relatively bulky (e.g. air tank, battery, etc.) are placed above the circular base, as close as possible to the center of the platform. Such compact placement ensures that the centre of gravity of the slider is placed close to its geometrical center. The base line dimension of the platform (approximately 350mm in diameter) is largely defined by the length of the air tank and the size of the air pressure regulators mounted over it. The components like thruster assembly and regulators are placed over the periphery of the circular base as shown in Fig.\ref{Blueprintmodel} that  provides maximum possible torque arm for controlling its motion. The a list of major components for building the slider platform is given in Table \ref{List_of_major_components}. A more elaborating design description is presented in the Table \ref{Blueprint_Detail_table1}, \ref{Blueprint_Detail_table2} of appendix section. Various components of the baseline structure used to construct the slider platform are built in laboratory using 3D printing technology which uses Polylactic Acid (PLA) \cite{drumright2000polylactic} as printing material. 
   \begin{table}[h]
\caption{List of Components}
\begin{center}
\begin{tabular}{|c|c|c|c|c|c|c|} 
\hline
Components         & Manufacturer & Product type/Specification   \\ \hline
Solenoid valve     & Festo        & MFH-2-M5                     \\ \hline
12V coil           & Festo        & MSFG-12-OD                   \\ \hline
Regulator          & Festo        & MS2-LR-QS6-D6-AR-BAR-B       \\ \hline
Airsoft regulator  & Polarstar    & MRS                          \\ \hline
Tank regulator     & Ninja        & HP UL Reg 4500psi            \\ \hline
Air tank           & DYE          & UL                           \\ \hline
Relay module       & Seeed        & Groove -2-Channel SPDT Relay \\ \hline
Upboard Computer            & Arduino      & Micro-controller\\ \hline
\end{tabular}
\end{center}
\label{List_of_major_components}
\end{table}

\begin{figure} 
\centerline{\includegraphics [width=0.8\textwidth] {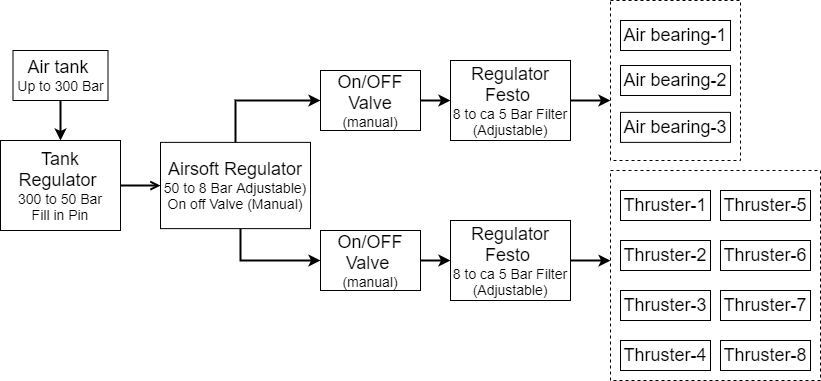}}
\caption{Schematic representation of Air-management system }\label{Air_management_fig}
\end{figure}
\subsection{Air Management System}
In order to drive all pneumatic components (primarily consist of air bearings and thruster assembly), the slider platform carries an air tank of size $1.2l$, which is filled with pressurized air and placed symmetrically about the $X_B-Z_B$ plane of the slider. The air tank structure can stores compressed air, pressurized up to $300$ bar. The air tank is connected with air-bearings (attached to bottom deck) and the thrusters through air tubes. A schematic representation of the air-management system is presented in Fig.\ref{Air_management_fig}. Pneumatic flow from air-tank is divided into two separate branches. Each of the branches is equipped with low-pressure regulator, which are capable to control the output pressure. The output air pressure, is regulated down to about $5$ bar for operation of air-bearings and about $7$ bar for thrusting mechanism. Relatively heavy regulators are mounted on the front side of the tank, to appropriately counteract the weight of the air tank along $Y_B-Z_B$ plane. The compressed air is evenly released through the porous surface of the air bearing in regulated manner. Three air bearings each of size $40$ mm are mounted in the bottom deck. While operational the three air-bearing together makes a plane of air cushion of a thickness of about $15$ microns, resulting in friction less motion of the platform over the friction-less table. 
\subsection{Actuation Mechanism}
\begin{figure}
\centering
\begin{minipage}{.5\textwidth}
\centerline{\includegraphics [width=0.5\textwidth] {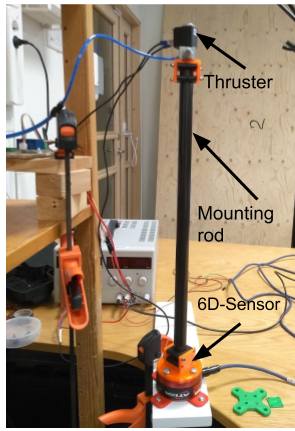}}
\caption{Experimental setup for thruster modeling}\label{Thruster_experimental_setup}
\end{minipage}%
\begin{minipage}{.5\textwidth}
  \centering
  \includegraphics[width=0.9\textwidth]{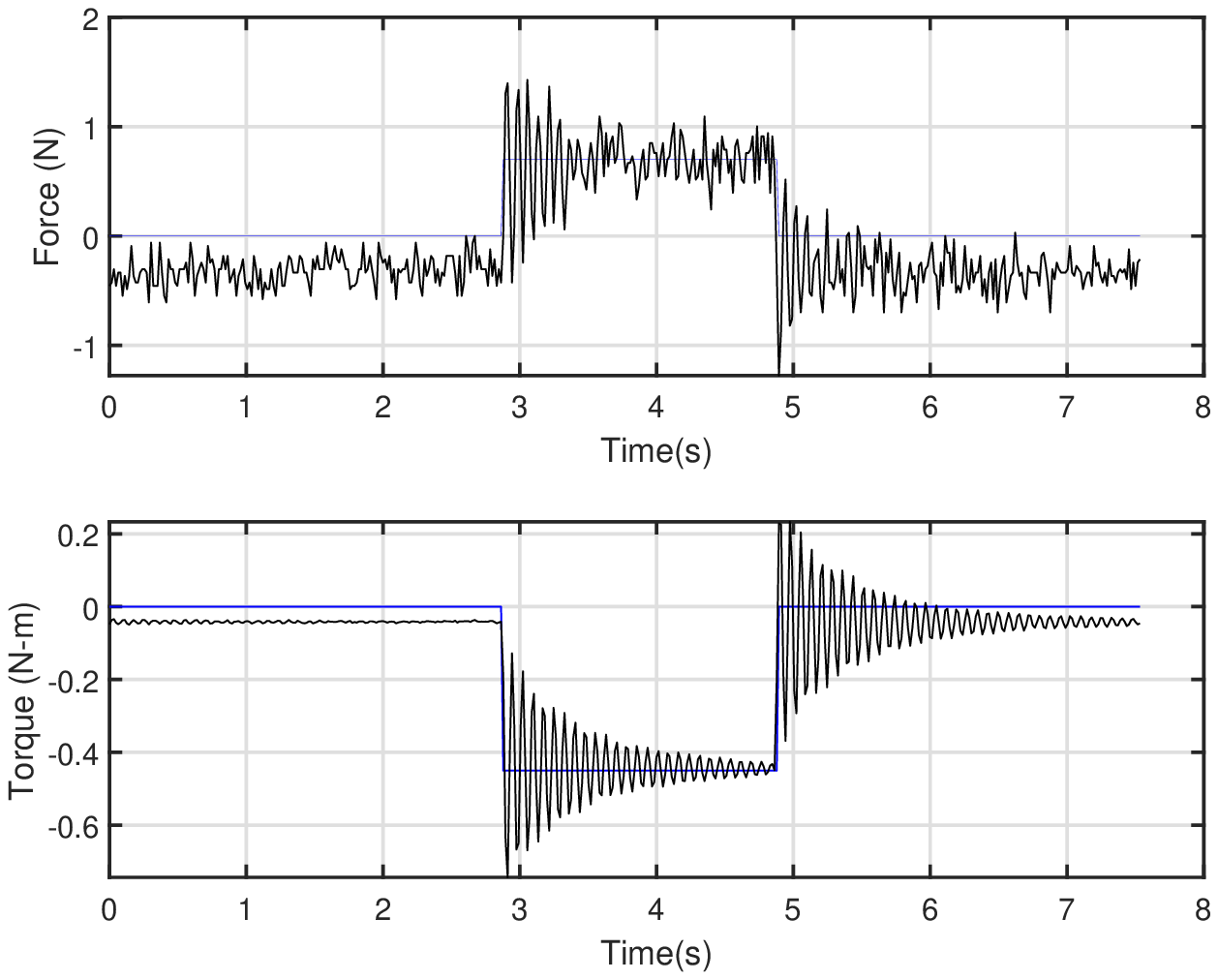}
  \caption{Excitation and measurement of output force and torque }\label{Thruster_output_model}
\end{minipage}
\end{figure}

In order to mobilize the slider platform in a controlled manner, it is equipped with eight small thrusters that operate in an on-off mode. The thrusters are synthesized with 3D printed nozzles integrated with $12$ V solenoid valves. The solenoid valves are controlled by relay modules, which operates on signals received from on-board computer or RC receiver through `Arduino' board. The energy is stored in $4S$ Lipo $1400$ mAh battery and it is regulated down to 12V for solenoid valves and down to $5$V for on-board computer. Engineering resin is used as material for synthesizing nozzles. Eight thrusters are distributed into four brackets, which are placed wide apart from the geometric centre, while maintaining a compact footprint of a square with a side about $39$mm. The placement of the thruster assembly is presented in Fig.\ref{Blueprintmodel}. Two thrusters sharing a bracket (One aligned with the $x$-axis and the other with the $y$-axis) maintain a small offset as shown in the Fig.\ref{Thruster_Placcement}. Such wide spread arrangement come up with large torque arms, which maximizes the magnitude of torque. Each of the thruster can produce a constant magnitude force of $0.7N$ while activated. The thruster assembly is capable to provide force and torque that is required to translate, as well to orient the slider over the friction-less table. An initial set of experiments has been cried out for construction of force and torque model of small thrusters. The experimental setup for thruster modeling is presented in Fig.\ref{Thruster_experimental_setup}. A thruster is mounted over a rod of length about $0.5$ m to provide a sufficiently large torque-arm. Such arrangement is required to mitigate the the resolution of the 6D Torque/Force sensor. A calibrated response of the thrusters are presented in Fig.\ref{Thruster_output_model}. It is evident that the delay in the thruster response are fairly insignificant for engineering practice. However, the undulation in measured signals are essentially due to presence of unwanted sensor noise.
Since the thrusters operates in a switching mode, a specific combination of thruster activation results into a particular type of directed motion of the slider. The specific combination of the thruster required to activate for various directed motions are presented in Table \ref{ThrustTable}. 
\begin{table}[ht] 
\centering
\caption{Thrust activation logic}
\label{ThrustTable} 
 \begin{tabular}{|c|c|c|c|c|c|c|c|c|}
\hline        
\textbf{Motion} & \multicolumn{8}{|c|}{\textbf{Thrusters}} \\ \hline
   & $T_{1}$ & $T_{2}$ & $T_{3}$ & $T_{4}$ & $T_{5}$ & $T_{6}$ & $T_{7}$ & $T_{8}$ \\ \hline
Forward           &          &          & \checkmark        &          & \checkmark         &          &          &          \\ \hline
Backward          & \checkmark        &          &          &          &           &          & \checkmark        &          \\ \hline
Left              &          & \checkmark        &          & \checkmark        &           &          &          &          \\ \hline
Right             &          &          &          &          &           & \checkmark        &          & \checkmark        \\ \hline
Clockwise         & \checkmark        &          &          &\checkmark        & \checkmark         &          &          & \checkmark        \\ \hline
C-Clockwise &          & \checkmark        & \checkmark        &          &           &\checkmark        & \checkmark        & \\
\hline        
\end{tabular}
  \end{table} 
  
\section{Initial Test} \label{initial_test}
An initial trial for the platform has been carried out in the Kiruna space lab facility. The experiment's goal is to test and prove the concept and demonstrate its manoeuvrability over the friction-less table. During the trial, the platform motion has been controlled manually using a remote control (RC) transmitter in an open-loop manner. The platform is equipped with an RC receiver, which is directly connected to the on-board computer. The on-board computer communicates with the thruster assembly and provides the `on/off' command state to a set of assigned thrusters.  The RC transmitter's sticks are set to be operated in an `on/off' fashion while Modulation and pulsation of the thrusters are controlled directly from the operator-end. The control logic for the RC transmitter has been designed as follows. Three dedicated channels of the RC transmitter are used to provide the command for the platform's directed platform motion. Among these three ones is assigned to control the forward and backwards movement,  while others are used for sidewise (towards left-right) and rotational (clockwise and counter-clockwise direction) motion. The actuation logic for selecting the set of the thruster for each directed motion is described in Table \ref{ThrustTable}.

During the initial trial, the Air tank has been filled with compressed air up to $200$ bar. The operating condition for the platform is set up as follows. The pressure regulator for the air bearings is set to $5$ bar, and the same for the thruster assembly is maintained at $7$ bar. With this setup, we reached about $7$ minutes of flight time. Airflow through the bearings is allowed continuously during the entire operation period whereas the thrusters are operated through the on-board computer (as commanded from the RC transmitter). The on-time for the thrusters is estimated to be $30\%$ of the entire flight time. Typically, the flight time is primarily dependent on the consumption rate of the pressurised air and hence on the thrusters usage. The battery life is over exceeding the lifetime of the air tank heavily; thus, it will not affect the platform's flight time. However, refiling of the air tank is required more often. A proof of demonstration of platform motion during the initial trial is recorded video-graphically and can be found in \cite{Youtube}.

During the initial trial, the platform motion is controlled manually and operated in open-loop mode. However, the realistic demonstration of various orbital manoeuvre and synchronised movement of multiple slider platform requires advanced control law to be operated autonomously in a closed-loop manner. The formulation of advanced model-based control design requires the  mathematical model of the platform. In view of that, the dynamic model of the slider platform consisting of the thruster based actuation mechanism is presented next. 
\section{Equation of Motion} \label{Mathematical_model}
In order to describe the equation of motion of the slider over friction-less platform, two frame references have been considered. An inertial frame of reference (denoted as $X-Y-Z$) is assumed to be attached on a corner point of the friction-less table. The axis of the inertial frames are considered to be directed along the length, width and height of the table. Another, moving frame of reference (denoted as $X_B-Y_B-Z_B$) is considered to be attached to the centre of gravity (CG) of the slider, which moves along the slider. The slider can translate over the table surface, as well it can rotate about its $Z_B$ axis. The position of the slider (denoted as $x,y$) is described in the inertial frame of reference. Since the various sensors and actuators are attached with the slider body, it is preferred to define its velocity and actuation forces in the body frame. Lets $v_x$ and $v_y$ denotes the velocity components of the slider expressed along  $X_B$ and $Y_B$ respectively. Since, the motion of the slider is restricted in $2-$D, the motion along $z$ component is ignored. Slider transnational kinematic equations of motion are described as:

\begin{figure}
\centering
\begin{minipage}{.5\textwidth}
\centerline{\includegraphics [width=0.9\textwidth] {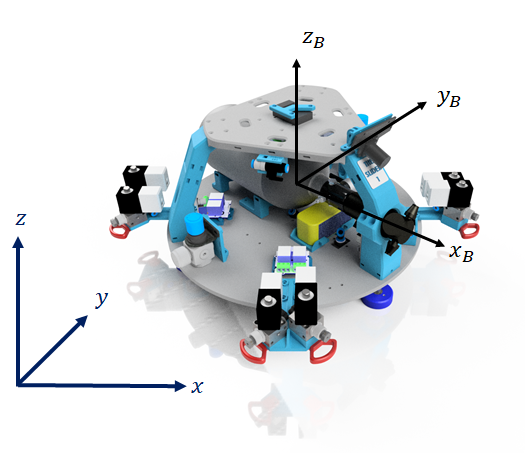}}
\caption{Reference frames used to describe the dynamics}\label{Ref_frame}
\end{minipage}%
\begin{minipage}{.5\textwidth}
  \centering
  \includegraphics[width=0.7\textwidth]{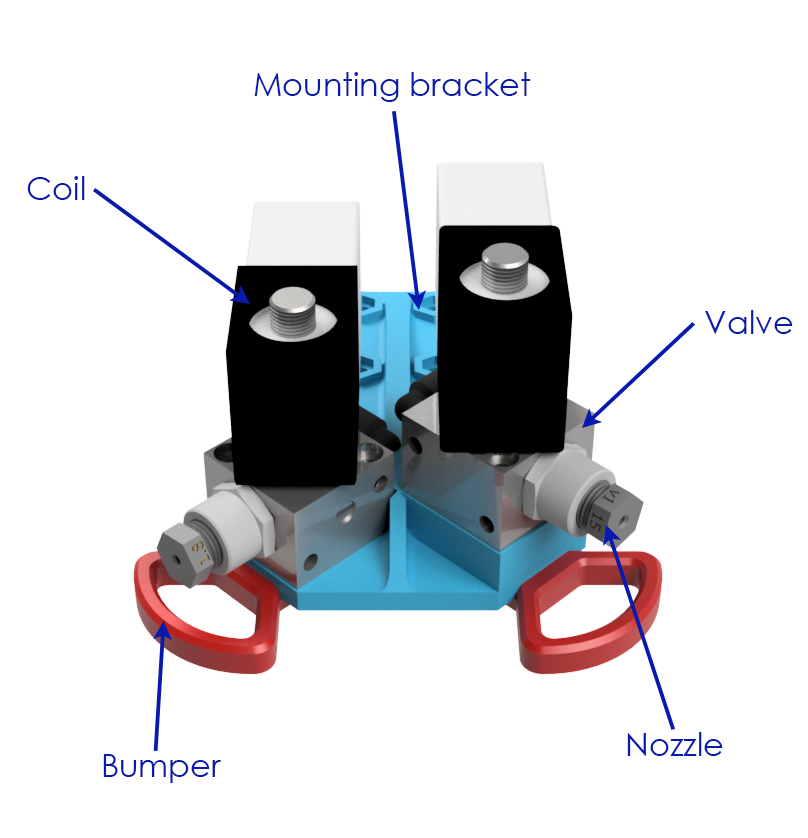}
  \caption{Thruster placement in a bracket }\label{Thruster_Placcement}
\label{schematic_Platform}
\end{minipage}
\end{figure}

\begin{equation} \label{Kinmatics}
    \left[ \begin{matrix}
   {\dot{x}}  \\
   {\dot{y}}  \\
\end{matrix} \right]=R_{B}^{I}\left[ \begin{matrix}
   {{v}_{x}}  \\
   {{v}_{y}}  \\
\end{matrix} \right]
\end{equation}
where, $R_{B}^{I}=\left[ \begin{matrix}
   \cos \theta  & -\sin \theta   \\
   \sin \theta  & \cos \theta   \\
\end{matrix} \right]$
represents the rotation matrix, that transform a vector from the body frame to inertial frame of reference. Here, $\theta$ represents the orientation of the slider (heading angle between $X$ and $X_B$ axis). Since the slider can rotate only about its $Z_B$ axis, it rotational velocity denoted as $r$ is directed along $Z_B$. Based on the fundamental Newton's law of motion \cite{junkins2009analytical}, transnational dynamics of the slider is formulated as
 \begin{equation}\label{dynamics}
    \left[ \begin{matrix}
   {{{\dot{v}}}_{x}}  \\
   {{{\dot{v}}}_{y}}  
\end{matrix} \right]=\left[ \begin{matrix}
   r{{v}_{y}}+\frac{{{f}_{x}}}{m}  \\
   -r{{v}_{x}}+\frac{{{f}_{y}}}{m}  
\end{matrix} \right]
\end{equation}
where $f_x,f_y$ denotes the actuation forces and $m$ represents the mass of the slider. Note that, the transnational dynamics incorporates the coriolis effects ($r{{v}_{x}},r{{v}_{y}}$) due to its rotational motion.
The rotational motion of the slider is formulated based on conservation of angular momentum \cite{sidi1997spacecraft} and presented as follows
\begin{equation}\label{rotation}
\left[ \begin{matrix}
   {\dot{\theta }}  \\
   {\dot{r}}  \\
\end{matrix} \right]=\left[ \begin{matrix}
   r  \\
   \frac{\tau }{{{I}_{zz}}}  \\
\end{matrix} \right]
\end{equation}
where, $\tau$ denotes the applied torque and $I_{zz}$ indicates the principal moment of inertia along the $Z_B$ direction. Note that the slider platform is designed in a balanced manner such that off diagonal components of moment of inertia matrix are negligible. combining the Eqs.(\ref{Kinmatics})-(\ref{rotation}), the dynamical equation of motion in compact form is represented as
 \begin{equation} \label{Dynamic_model}
   \left[ \begin{matrix}
   {\dot{x}}  \\
   {\dot{y}}  \\
   {\dot{\theta }}  \\
   {{{\dot{v}}}_{x}}  \\
   {{{\dot{v}}}_{y}}  \\
   {\dot{r}}  \\
\end{matrix} \right]=\left[ \begin{matrix}
   {{v}_{x}}\cos \theta -{{v}_{y}}\sin \theta   \\
   {{v}_{x}}\sin \theta + {{v}_{y}}\cos \theta   \\
   r  \\
   r{{v}_{y}}+\frac{{{f}_{x}}}{m}  \\
   -r{{v}_{x}}+\frac{{{f}_{y}}}{m}  \\
   \frac{\tau }{{{I}_{zz}}}  \\
\end{matrix} \right]
\end{equation}
The slider's actuation unit is equipped with a total number of eight  small thrusters attached with the platform. The control action, i.e. forces and torque components are related with the actuation of thruster units, modeled as
\begin{equation}\label{Force_thrust_relation}
\left[ \begin{matrix}
   {{f}_{x}}  \\
   {{f}_{y}}  \\
   \tau   \\
\end{matrix} \right]=\left[ \begin{matrix}
  \sum\limits_{k=1}^{8}{{{T}_{k}}\cos {{\beta }_{k}}}  \\
  \sum\limits_{k=1}^{8}{{{T}_{k}}\sin {{\beta }_{k}}}  \\
   \left( \sum\limits_{k=1}^{8}{\left( {{T}_{k}}r_{{{T}_{k}}}^{y}\cos \beta_k -{{T}_{k}}r_{{{T}_{k}}}^{x}\cos {{\beta }_{k}} \right)} \right)  \\
\end{matrix} \right]
\end{equation}
where, $T_k$ denotes the constant thrust magnitude,  $(r_{{{T}_{k}}}^{x},r_{{{T}_{k}}}^{y})$ together indicates the position of the $k^{th}$ thruster in $X_B,Y_B$ plane and ${{\beta }_{k}}$ represents its orientation with respect to $X_B$ axis.

\begin{table}[]
\caption{Numerical values for system and simulation parameters}
\label{Parameter_Table}
\begin{subtable}{.5\linewidth}
      \centering
\caption{Position and orientation of individual thruster}
\begin{tabular}{|c|c|c|}
\hline
Thruster & \begin{tabular}[c]{@{}c@{}}Position ($r_{{{T}_{k}}}^{x},r_{{{T}_{k}}}^{y}$)\\  (mm)\end{tabular} & \begin{tabular}[c]{@{}c@{}}Orientation $\beta_k$\\ (deg)\end{tabular} \\ \hline
$T_1$    & $(195,-140)$                                                                                     & $0$                                                                   \\ \hline
$T_2$    & $(140, -195)$                                                                                    & $270$                                                                 \\ \hline
$T_3$    & $(-195, -140)$                                                                                   & $180$                                                                 \\ \hline
$T_4$    & $(-140, -195)$                                                                                   & $270$                                                                 \\ \hline
$T_5$    & $(-195, 140)$                                                                                    & $180$                                                                 \\ \hline
$T_6$    & $(-140, 195)$                                                                                    & $90$                                                                  \\ \hline
$T_7$    & $(195, -140)$                                                                                    & $0$                                                                   \\ \hline
$T_8$    & $(140, 195)$                                                                                     & $90$                                                                  \\ \hline
\end{tabular}
\label{Thruster_Position_orientation}
    \end{subtable}%
\begin{subtable}{.5\linewidth}
      \caption{System parameters}
      \centering
\begin{tabular}{|c|c|}
\hline
Parameters                                                             & Values          \\ \hline
Mass ($m$)                                                             & $4.436$ kg      \\ \hline
Moment of inertia ($I_{zz}$)                                             & $1.092$ $kg-m^2$ \\ \hline
\begin{tabular}[c]{@{}c@{}}Control command \\ time step\end{tabular}   & $0.5$ s           \\ \hline
\begin{tabular}[c]{@{}c@{}}System Propagation\\ time step\end{tabular} & $0.01$ s        \\ \hline
\begin{tabular}[c]{@{}c@{}}Minimum\\ on time of thruster\end{tabular}  & $10$ ms         \\ \hline
$T_{lb}, T_{ub}$                                                       & $0, 0.7$ N      \\ \hline
\end{tabular}
  \end{subtable}%
\end{table}
\section{Model based Open Loop Simulation} \label{Simulation_Results}
In order to validate the mathematical model, an open-loop simulation has been carried out in this section. Various numerical parameters used for the simulation study is presented in Table \ref{Parameter_Table}. Initially the slider platform is considered to be resting at the corner of the table, i.e. $[x,y,\theta,v_x,v_y,r]=0_{6 \times 1}$. It is intended to excite the platform model with two-step ramp type input command (combining both the positive and negative actuation). The open-loop control excitation signal is depicted in the fourth subplot of Fig.\ref{Simulation_velocity_input}. The identical control input is considered to be applied along the three channels, i.e. $f_x,f_y$ and $\tau$. The simulation-based validation has been carried out in the two-step approach. In the first step, the ideal system response is evaluated by propagating of the dynamic model  Eq.(\ref{Dynamic_model}) in the presence of continuous-time force and torque command. Here, the realistic actuator model consisting of eight thruster assembly, is ignored. Ideal time response of the slider pose and velocities due to continuous input command are presented in Figs. \ref{Simulation_pose} and \ref{Simulation_velocity_input} respectively. Here the blue dotted lines represent the ideal system responses. In the next step, the realistic actuator model with its physical limitations (such as maximum thrust bound, minimum on-time etc.) is accounted for in the simulation study. The open-loop force and torque command, i.e. $f_x,f_y,\tau$ needs to be realised by actuating the on-off thrusters assembly, i.e. $T_1 \cdot T_8$. Rewriting the Eq.(\ref{Force_thrust_relation}), the relation between $f_x,f_y,\tau$ and $T_1 \cdot T_8$ can be expressed as 
\begin{equation} \label{constraint}
    Ax=b
\end{equation}
where, $b=\left[\begin{array}{c}
f_{x} \\
f_{y} \\
\tau
\end{array}\right],x=\left[\begin{array}{c}
T_{1} \\
T_{2} \\
\vdots \\
T_{8}
\end{array}\right], A =
\left[\begin{array}{ccc}
\cos \beta_{1} &  & \cos \beta_{8} \\
\sin \beta_{1} & \vdots & \sin \beta_{8} \\
{\left(r_{{{T}_{1}}}^{y}\cos \beta_1 -r_{{{T}_{1}}}^{x}\cos {{\beta }_{1}} \right)}   &  & {\left(r_{{{T}_{8}}}^{y}\cos \beta_8 -r_{{{T}_{8}}}^{x}\cos {{\beta }_{8}} \right)}
\end{array}\right]$
\\
Note that the matrix $A$ is known and constant, determined by placement of each thruster's and its orientation. 

\begin{figure} [ht]
\centerline{\includegraphics [width=0.8\textwidth] {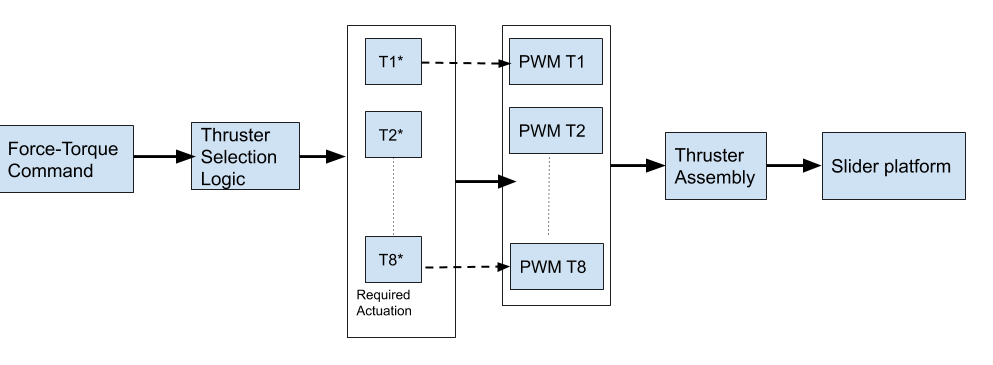}}
\caption{Schematic representation of low level actuation logic}
\label{Thruster_activation_module}
\end{figure}
\subsection*{Thruster Selection Logic} The objective of the thruster selection logic is to determine the set of thrusters required to be activated in order to achieve the desired commanded force and torque inputs. A pseudo-inverse based solution of Eq. (\ref{constraint}) is adopted in \cite{ghobadi2020reliability}. The method incorporates a null space adjustment depending on the sign of input command. In this paper, the above problem has been translated into an optimization framework \cite{martel2004optimal} as follows:

\begin{align}
&  Min J = x^Tx\\
       & S.T.  Ax=b\\
       & x_{ub} \ge x \ge x_{lb}
\end{align}
\begin{figure}
\centering
\begin{minipage}{.5\textwidth}
\centerline{\includegraphics [width=1\textwidth] {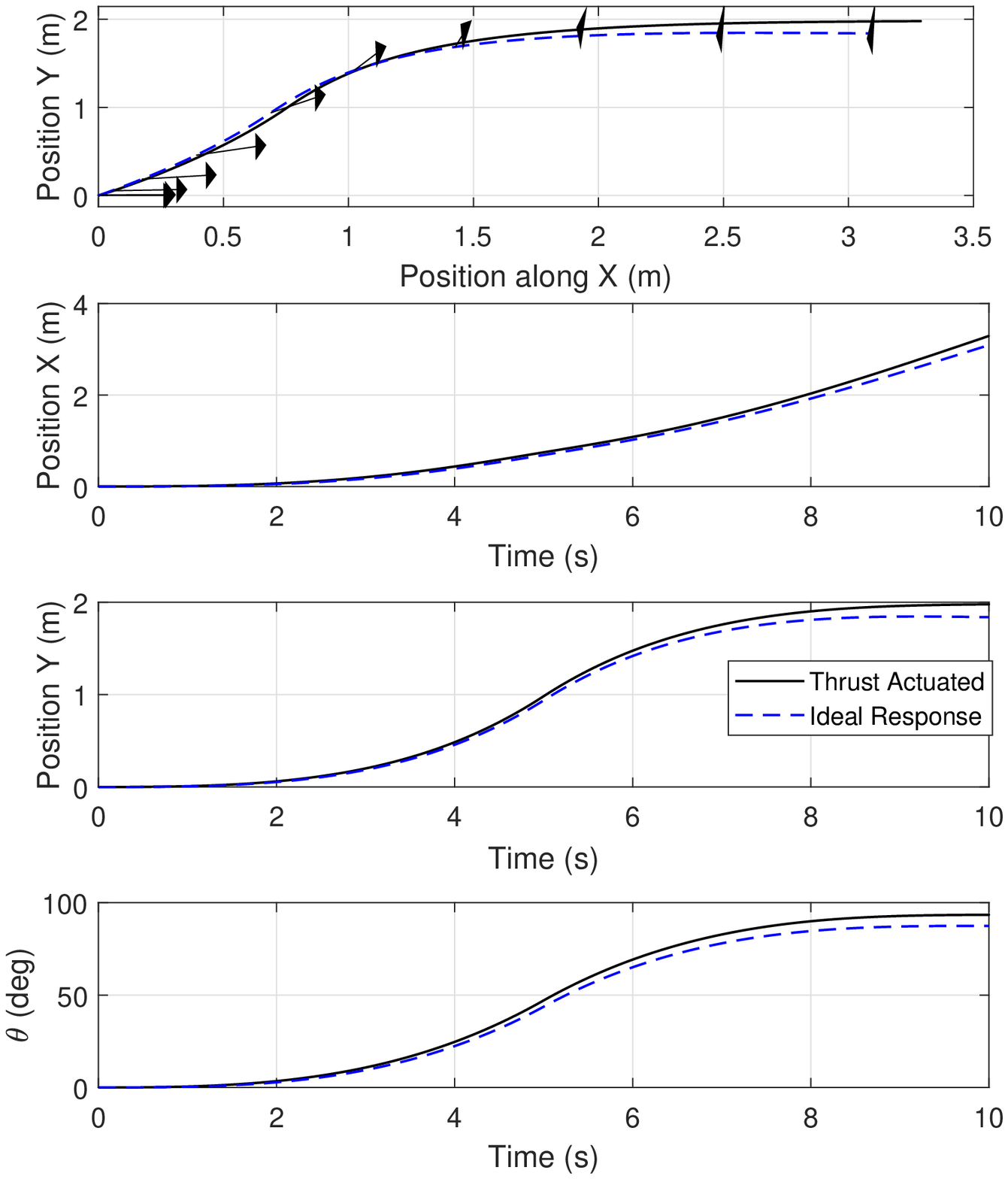}}
\caption{Open-loop response of slider pose}\label{Simulation_pose}
\end{minipage}%
\begin{minipage}{.5\textwidth}
  \centering
  \includegraphics[width=1\textwidth]{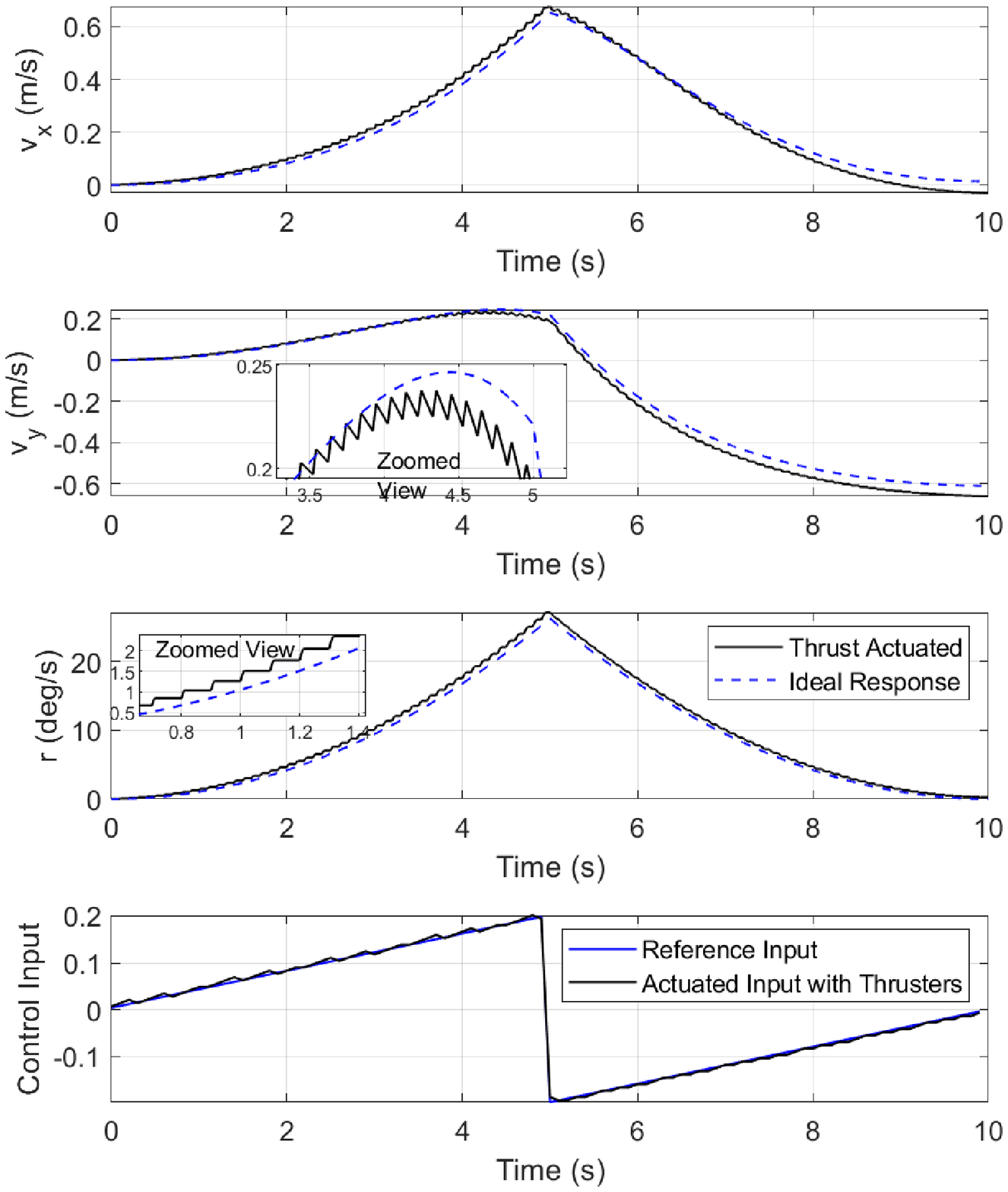}
  \caption{Variation of slider velocity profiles and input excitation }\label{Simulation_velocity_input}
\end{minipage}
\end{figure} 
The solution of the above optimization problem provides the optimal magnitude of each thruster ($T_k^*$) that collectively mitigates the necessary force and torque demand, while minimizes the actuation effort and ensuring the physical limitations of each thruster. Note that, the magnitude of each thruster $T_k^*$ obtained based on thruster selection logic can be any value between $x_b$ and $u_b$ limits. However, in practice these thrusters operate in an ON-OFF mode, i.e. either it can provide a full actuation with a maximum thrust or operates in no thrust mode. In order to address this, the pulse width modulation (PWM) technique is incorporated. Dedicated PWM units are assigned for each thruster, as shown in Fig.\ref{Thruster_activation_module}. The PWM block generates a sequence of pulses in such a way that the average thrust produced by each thruster closely follows the required magnitude $T_k^*$. 

\begin{figure}
\centering
\begin{minipage}{.5\textwidth}
\centerline{\includegraphics [width=1\textwidth] {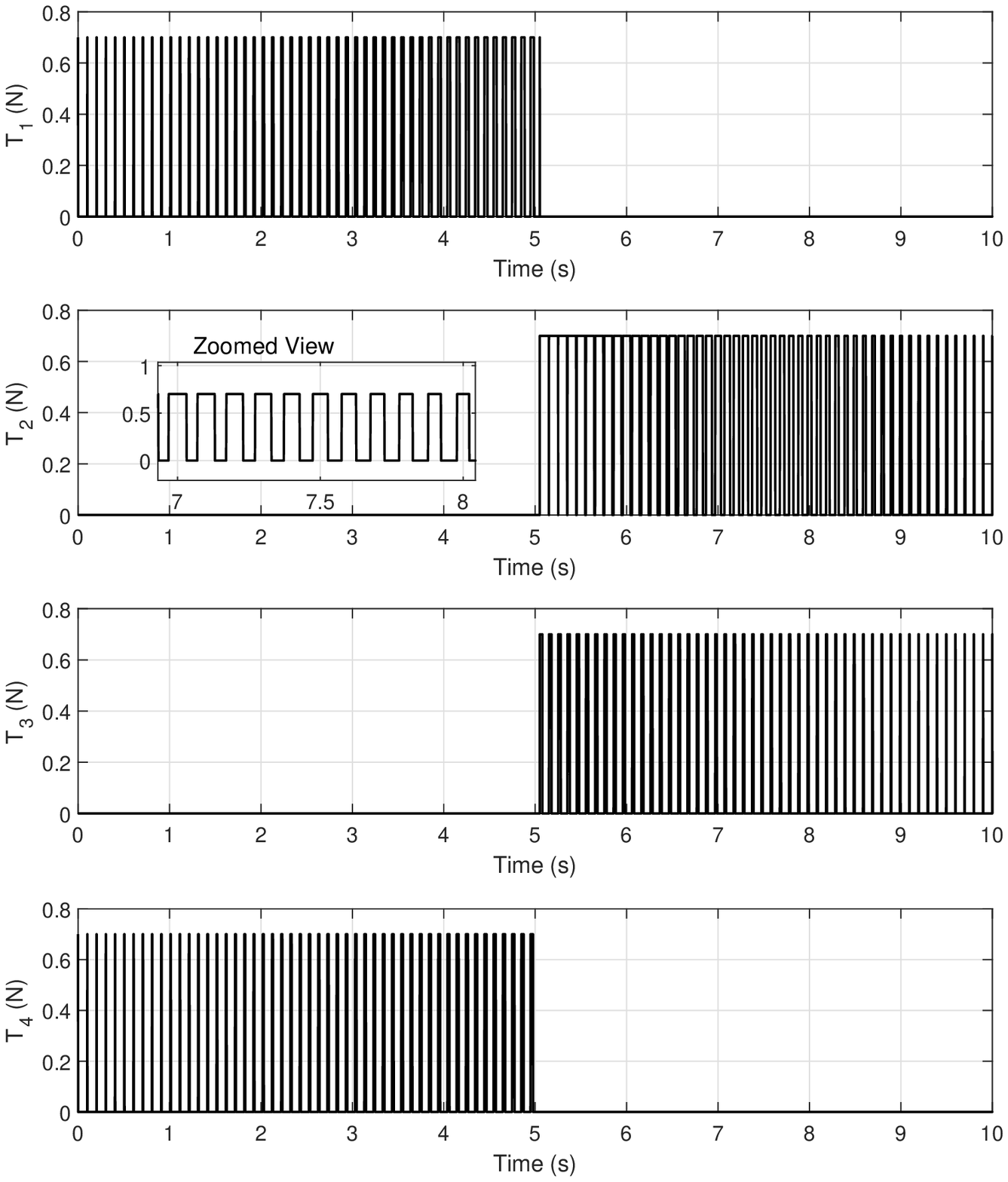}}
\caption{Actuation command for thrusters $T_1-T_4$}\label{T1_T4}
\end{minipage}%
\begin{minipage}{.5\textwidth}
  \centering
  \includegraphics[width=1\textwidth]{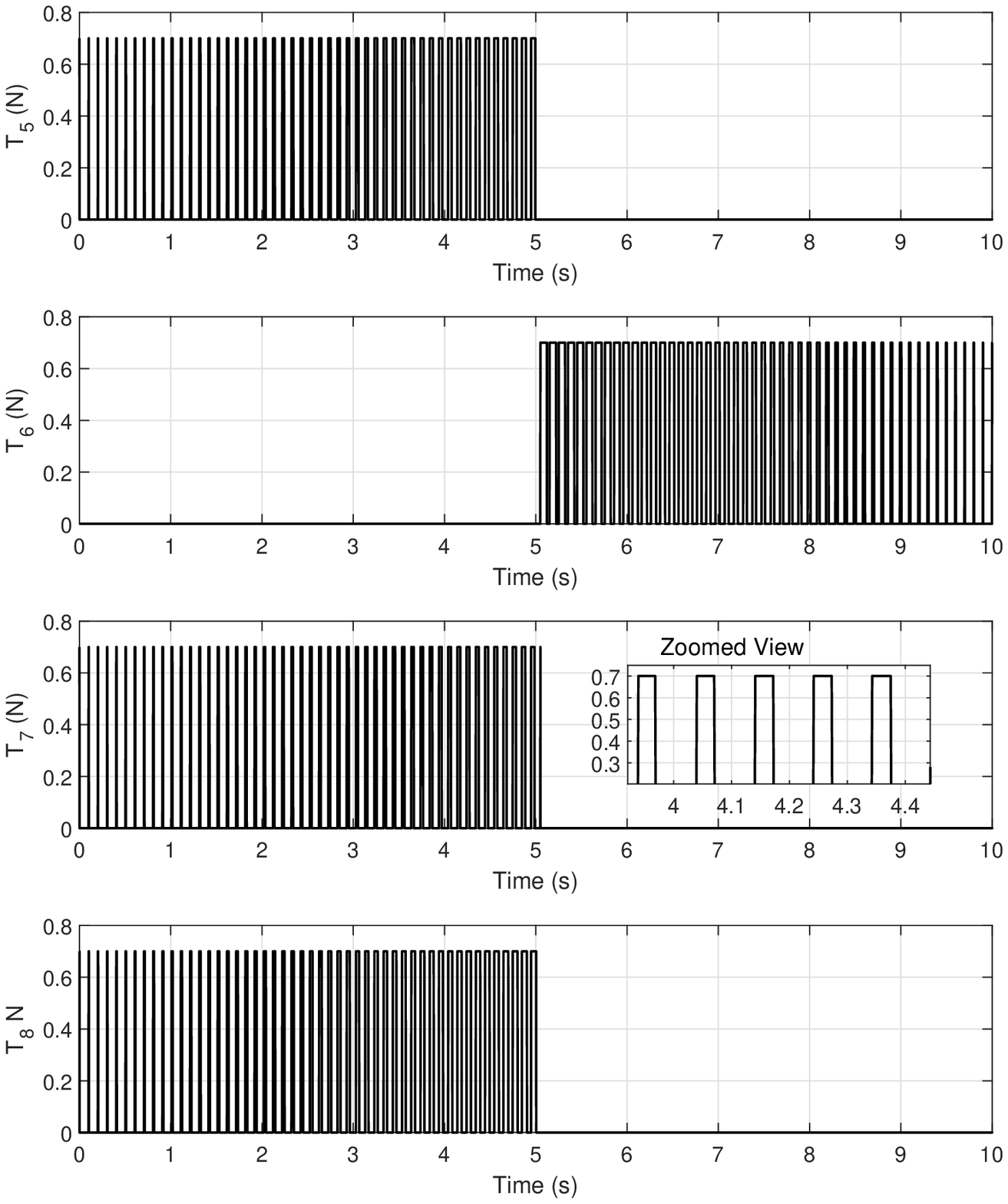}
  \caption{Actuation command for thrusters $T_5-T_8$ }
\label{T5_T8}
\end{minipage}
\end{figure}

The implementation of actuator based simulation has been carried out as follows. The input commands ($f_x,f_y$ and $\tau$) are processed through the zero-order-hold mechanism and its magnitude is hold at a constant value for each control time step, i.e. $0.5$ s. During this period, the input signal is processed through various intermediate blocks, as presented in Fig. \ref{Thruster_activation_module}. Each PWM blocks are set to be operated with a frequency of $10$ Hz. The resulting sequence of output pulses for each thruster $T_1-T_8$ are presented in Figs.\ref{T1_T4} and \ref{T5_T8}. With the application of pulse input as actuation command for thrusters, the nonlinear dynamic model is propagated, and the corresponding time responses are presented in Figs.\ref{Simulation_pose} and \ref{Simulation_velocity_input}. Here the solid black lines represent the thrust actuated system responses. In the first subplot of Fig. \ref{Simulation_pose}, the slider trajectory is presented where the arrowheads are indicating its orientation at the given point. Detailed orientation profile is presented in the fourth subplot of Fig.\ref{Simulation_pose}. It is evident that the thruster based actuation mechanism closely resembles the ideal responses. However, the zoomed view in Fig. \ref{Simulation_velocity_input} shows a non-smooth behaviour, which is due to pulsating mode of actuation input.    

\section{Conclusion} \label{Conclusion}
In this article, a friction-less floating robotic test-bed facility for a hardware-in-loop experimental study of a planar satellite that has indigenously developed at Lule\aa\ University of Technology is presented. The details design description of the physical platform, with the each component specification was presented. Moreover a mathematical model describing its dynamic motion was formulated. This capability can be used to conduct research on coordinated control of spacecraft teams. A successful initial test has been conducted in a open manually operated loop remote control manner. Multiple simulation studies has been carried out with formulated mathematical model in various scenarios. An optimization based actuator allocation logic has been developed and tested in simulation framework. The on-board actuators, composed of eight thrusters, which is equivalent to the actuators configuration of typical spacecraft and further establishes close realistic equivalence. Additionally, multiple such robotic platform can be operated simultaneously over the friction less table. This state-of-the-art dynamic hardware-in-the-loop emulation facility will continue to be fruitful for the advancement of spacecraft proximity operations research.

\bibliographystyle{./bibliography/IEEEtran.bst}
\bibliography{mybib.bib}
\section{appendix}
\setcounter{table}{0}
\setcounter{figure}{0}    
\renewcommand{\thetable}{A\arabic{table}}
\renewcommand\thefigure{A\arabic{figure}}
Additional supplementary details of each component used to construct the slider platform are presented in this section.
Each component is chronologically identified with a serial number between $1-55$. The location of each component is indicated with the help of the blueprint model of the slider platform, as shown in Fig. \ref{Blueprint_Detail}. Various description such as product specification, materiel used for fabrication etc. are listed for each components in Tables \ref{Blueprint_Detail_table1} and \ref{Blueprint_Detail_table2}. A more detail design related supplementary materials of the slider platform are made available in the GitHub repository \cite{github}, which include 3D CAD model, data for 3D printing, laser cutting, blueprint diagram and extensive list of various components. 

\begin{figure} [H]
\centering
\begin{subfigure}[b]{1\textwidth}
    \includegraphics[width=0.8\textwidth]{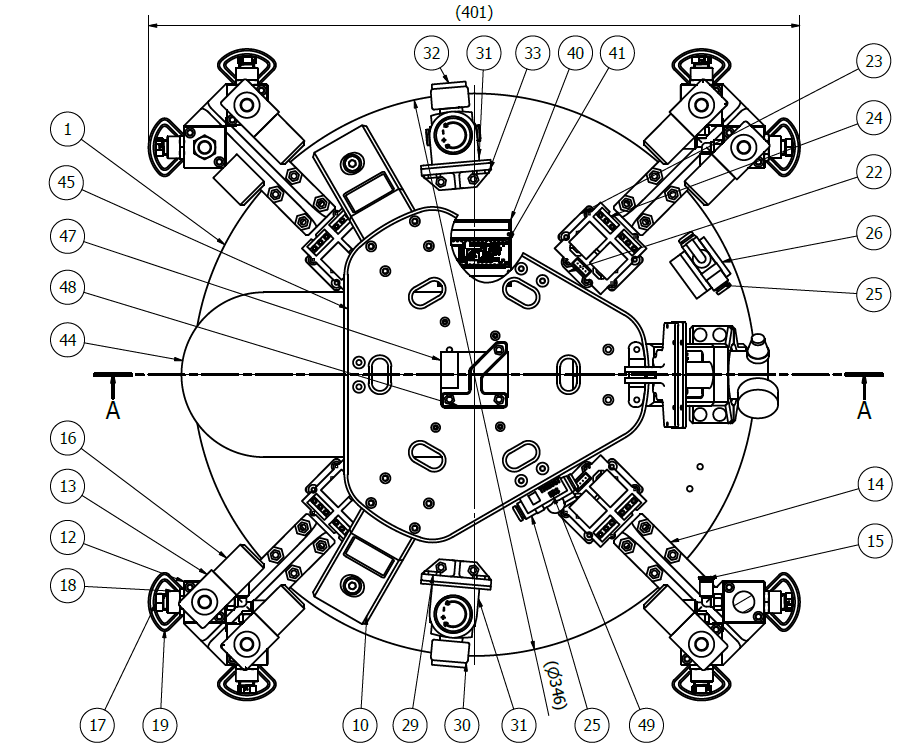}
    \label{Blueprint_Detail_topview}
    \caption{Top view}
\end{subfigure}    
    
\begin{subfigure}[b]{1\textwidth}
   \label{Blueprint_Detail_sideview}
  \includegraphics[width=0.8\textwidth]{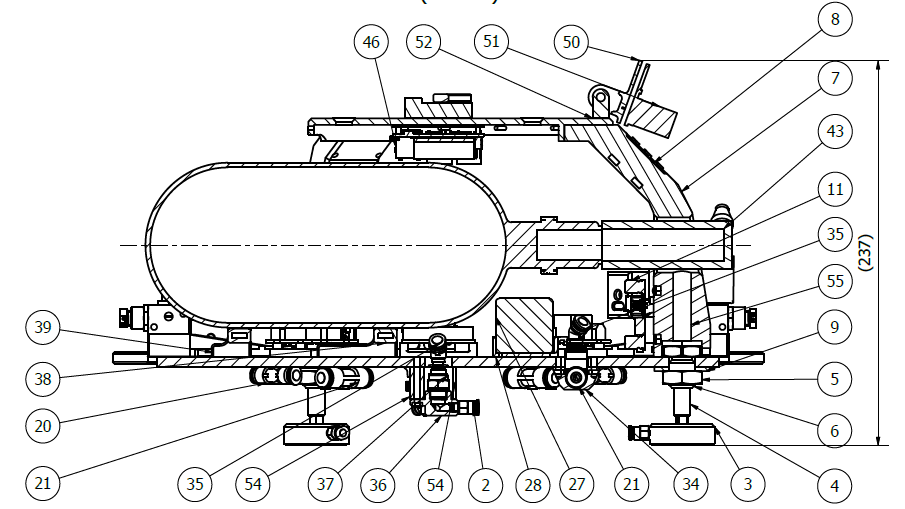}
  \caption{Side view}
  \end{subfigure}
\caption{Blueprint model of slider platform indicating detailed components}
\label{Blueprint_Detail}
\end{figure}

\begin{table}[H] 
\caption{List of Components (serial No. 1- 25) indicated in Figs \ref{Blueprint_Detail}}
\begin{center}
\begin{tabular}{ | l | l | l | l | l | }
\hline
	ITEM & No. of QTY & Product type/Specification & Material & Description \\ \hline
	1 & 1 & Base-6mm & Polycarbonate, Clear & Laser Cut (6mm) \\ \hline
	2 & 6 & 130778 QSM-M5-4-100 &  & QSM-push-in fitting \\ \hline
	3 & 3 & S104001 &  & Air bearing \\ \hline
	4 & 3 & S8013B11 & Stainless Steel & Air bearing bolt \\ \hline
	5 & 3 & S8013H04-NuT & Stainless Steel & Air bearing Hex nut \\ \hline
	6 & 3 & S8013H04-ScrewNut & Brass, Soft Yellow & Air bearing housing \\ \hline
	7 & 1 & Low-top-platform-mount-front-V4-1st part & PLA & 3D print \\ \hline
	8 & 1 & Name-plate & PLA & 3D print \\ \hline
	9 & 3 & Leg-washer & PLA & 3D print \\ \hline
	10 & 2 & Low-top-platform-mount-v2 & PLA & 3D print \\ \hline
	11 & 3 & LM2596 DC-DC StepDown Converter v1 &  & Step-down voltage regulator \\ \hline
	12 & 8 & 4573 MFH-2-M5 &  & MFH-Solenoid valve \\ \hline
	13 & 8 & 320410 MSFG-12-OD---(P) &  & MSFG-p-Solenoid coil \\ \hline
	14 & 2 & Valve-holder-v3 & PLA & 3D print \\ \hline
	15 & 6 & 153333 QSML-M5-4 &  & QSML-Push-in L-fitting \\ \hline
	16 & 8 & Festo-connector-cover & PLA & 3D print \\ \hline
	17 & 8 & Nozzle-SLA-base & Resin "Grey pro"  & 3D print (SLA) \\ \hline
	18 & 8 & 8030314 NPFC-R-G18-M5-FM &  & NPFC-R-Threaded fittings \\ \hline
	19 & 8 & Nozzle-bumper-V2 & TPU 95A & 3D print \\ \hline
	20 & 4 & 153374 QSMY-6-4 &  & QSMY-Push-in Y-connector \\ \hline
	21 & 5 & 153129 QST-6 &  & QST-Push-in T connector \\ \hline
	22 & 4 & T-piece-clamp & PLA & 3D print \\ \hline
	23 & 4 & Relay-mount & PLA & 3D print \\ \hline
	24 & 4 & Grove 2 Channel SPDT Relay &  & Relay \\ \hline
	25 & 2 & 153484 QH-QS-6 &  & QH-QS-ball valve \\ \hline
\end{tabular}
\end{center}
\label{Blueprint_Detail_table1}
\end{table}

\begin{table}[H] 
\caption{List of Components (serial No. 26- 55) indicated in Figs \ref{Blueprint_Detail_table2}}
\begin{center}
\begin{tabular}{ | l | l | l | l | l | }
\hline
	ITEM & No. of QTY & Product type/Specification & Material & Description \\ \hline
	26 & 1 & Valve holder-V3 & PLA &  \\ \hline
	27 & 1 & Zippy compact 1400 &  & Battery \\ \hline
	28 & 1 & Battery-case-zippy-1400i-V2 & PLA & 3D print \\ \hline
	29 & 1 & Regulator-mount-v2 & PLA & 3D print \\ \hline
	30 & 1 & 8086628 MS2-LR-M5-D6-AR-BAR-B &  & Regulator w filter Air bearings \\ \hline
	31 & 2 & 5003640 MS2-LR/LFR-B &  & MS2-WR (p)-Mounting bracket \\ \hline
	32 & 1 & 8086644 MS2-LFR-QS6-D6-AR-BAR-C-M-B &  & Regulator Thrusters \\ \hline
	33 & 1 & Regulator-mount-long-v3 & PLA & PLA 3D print \\ \hline
	34 & 1 & T-piece-holder & PLA & 3D print \\ \hline
	35 & 2 & 130618 QSW-6HL &  & QSW-HL-Push-in connector \\ \hline
	36 & 1 & Splice-sla & Resin "Grey pro"  & 3D print (SLA) \\ \hline
	37 & 1 & 186096 QS-G1/8-6 &  & QS\_G-Push-in fitting \\ \hline
	38 & 1 & Bottle mount V3-extended & PLA & 3D print \\ \hline
	39 & 1 & Bottle mount V3 & PLA & 3D print \\ \hline
	40 & 1 & Arduino shield &  & Arduino shield \\ \hline
	41 & 1 & Arduino Nano 33 IoT &  & Arduino \\ \hline
	42 & 1 & Arduino-w-shiled-holder & Resin "Black" & 3D print (SLA) \\ \hline
	43 & 1 & Polarstar-regulator &  & Airsoft regulator \\ \hline
	44 & 1 & paintball air tank-DYE &  & Paintball tank 1,1l \\ \hline
	45 & 1 & top ring-6MM-V2 & PLA & 3D print \\ \hline
	46 & 1 & UP\_BOARD\_\&\_HEAT\_SINK\_by\_JMJV &  & Up-boad computer \\ \hline
	47 & 1 & Futaba-reciever & PLA & RC reciever \\ \hline
	48 & 1 & Futaba-controller-hodler & PLA & 3D print \\ \hline
	49 & 1 & Valve holder-V4 & PLA & 3D print \\ \hline
	50 & 1 & holder\_ps & PLA & 3D print \\ \hline
	51 & 1 & ps-camera &  & Playstation camera \\ \hline
	52 & 1 & Gopro-mount & PLA & 3D print \\ \hline
	53 & 2 & Valve-holder-v3-mirror & PLA & 3D print \\ \hline
	54 & 1 & Spacer & PLA & 3D print \\ \hline
	55 & 1 & Low-top-platform-mount-front-V4-2nd part & PLA & 3D print \\ 
\hline        
\end{tabular}
\end{center}
\label{Blueprint_Detail_table2}
\end{table}

\end{document}